\pdfoutput=1

\documentclass[11pt]{article}

\usepackage{acl}

\usepackage{times}
\usepackage{latexsym}

\usepackage[T1]{fontenc}

\usepackage[utf8]{inputenc}

\usepackage{microtype}

\usepackage{booktabs} %
\usepackage{graphicx} %
\usepackage{pgfplots} %
\usepackage{soul} %
\usepackage{listings} %
\usepackage{fdsymbol} %
\usepackage{cleveref} %
\usepackage{hyperref}
\usepackage{tikz}
\usepackage{annotation}

\pgfplotsset{compat=1.17}

\title{
    \vspace*{-0.5in}
    {\raggedright\small ACL 2023\\[0.25in]}
    The Elephant in the Room: Analyzing the Presence of Big Tech\\in Natural Language Processing Research
}

\author{Mohamed Abdalla${^\spadesuit}$\thanks{\hspace{1.5mm}Equal contribution.} , Jan Philip Wahle${^\clubsuit}$\textsuperscript{*}\\
{\bf Terry Ruas${^\clubsuit}$, Aurélie Névéol$^{\vardiamondsuit}$, Fanny Ducel$^{\diamondsuit}$, Saif M. Mohammad$^{\Phi}$, Karën Fort$^{\diamondsuit}$}\\
$^{\spadesuit}$Institute for Better Health, Canada, $^{\clubsuit}$University of Göttingen, Germany\\
$^{\vardiamondsuit}$Université Paris-Saclay, CNRS, LISN, France, $^{\diamondsuit}$Sorbonne Université / LORIA, France\\
$^{\Phi}$National Research Council, Canada\\
\texttt{msa@cs.toronto.edu} \hspace{3mm} \texttt{wahle@uni-goettingen.de}\\}

\begin{document}
\maketitle
\AddAnnotationRef

\begin{abstract}
Recent advances in deep learning methods for natural language processing (NLP) have created new business opportunities and made NLP research critical for industry development. As one of the big players in the field of NLP, together with governments and universities, it is important to track the influence of industry on research. In this study, we seek to quantify and characterize industry presence in the NLP community over time. Using a corpus with comprehensive metadata of 78,187 NLP publications and 701 resumes of NLP publication authors, we explore the industry presence in the field since the early 90s. We find that industry presence among NLP authors has been steady before a steep increase over the past five years (180\% growth from 2017 to 2022). A few companies account for most of the publications and provide funding to academic researchers through grants and internships. Our study shows that the presence and impact of the industry on natural language processing research are significant and fast-growing. This work calls for increased transparency of industry influence in the field.
\end{abstract}

\section{Introduction}

\noindent Research is influenced by several entities such as academia, government, and industry.
Their roles and degrees of influence change over time. 
Recent deep learning advances in natural language processing (NLP) have created a spurt of new business opportunities, making NLP research critical for industry development.
In turn, we are observing a greater presence of large technology companies (Big Tech) on NLP research than ever before.
This influence on research can be beneficial. Companies provide funding, and participate in open science initiatives \cite{gulbrandsen2005industry, hottenrott2011industry}. However, there are growing voices of concern about scientific independence and power \cite{AbdallaAIES21, whittaker2021steep} --- from controlling the accessibility of massive amounts of computing power to powerful language models \cite{devlin-etal-2019-bert, gpt3}.

\begin{figure}[t]
    \centering
   \includegraphics[width=\columnwidth]{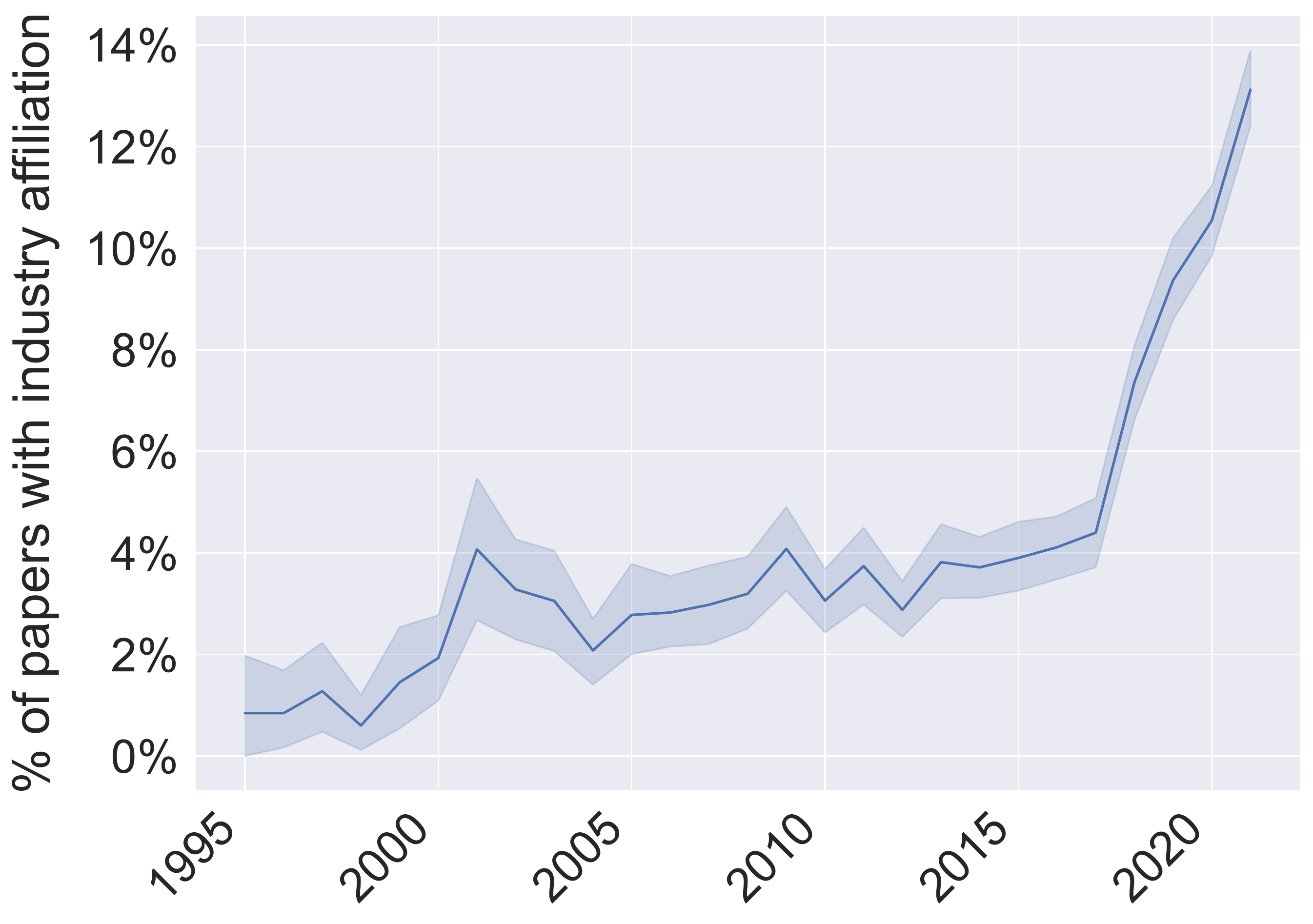}
    \caption{Proportion of papers in the ACL anthology with at least one industry affiliation (1995 - 2022).}
    \label{fig:overall-relative-presence}
    \vspace*{-3mm}
\end{figure}

It is in the interest of any community to pause and reflect on who has the power, what their interests and values might be, and how they are influencing research.
Thus, this recent period of great change in NLP research merits a reflection on the changing role of industry in research, as well as their implications on society at large.
Such a reflection will naturally require many broad approaches studying the socio-, economic-, and political forces at play in combination with technology.
Our work aims at facilitating such a reflection by quantifying and characterizing the presence of Big Tech in NLP research, especially during this time of change. %
As such, it puts a spotlight on the changing role of industry in NLP research.

\noindent \textbf{The changing role of industry in NLP research.} Our analysis of papers published in the ACL anthology with at least one company affiliation (Figure~\ref{fig:overall-relative-presence} - see~\Cref{sec:industry_who}) confirms that we are currently experiencing dramatic change.
As highlighted by the industry track at NAACL\footnote{\href{https://2022.naacl.org/calls/industry/}{https://2022.naacl.org/calls/industry/}}, industry research can offer a focus on applied research and technical challenges such as scaling robustness.  
By establishing large industrial research labs or through collaborations with academics, the industry has fostered increased attention to these research topics. 

\noindent \textbf{Favorable assessments.} Increased focus on NLP from industry can result in seemingly positive outcomes such as increased funding (e.g., through grants, awards, and scholarships). 
More generally, in the field of policy research, industry funding is largely viewed as positive~\cite{holman2021me}, with industry-funded research having been shown to result in more patents~\cite{hottenrott2011industry}, and more publications overall~\cite{gulbrandsen2005industry}. 

\noindent \textbf{Warnings.} However, in the philosophy of science, industry funding is largely viewed as ``corrosive to scientific inquiry''~\cite{holman2018promise,holman2021me}.
This view has been increasingly reflected in the work of AI ethics researchers who have highlighted concerns regarding the impact of such a large and growing industry presence~\cite{ahmed2020democratization,AbdallaAIES21,whittaker2021steep}. 
\citet{AbdallaAIES21} raise concerns about the lack of impartiality of research and undue influence on research questions. 
Others raise concerns about the centralization of resources, where only those with industry affiliations can research the latest language technologies~\cite{whittaker2021steep,ahmed2020democratization}. 
There is also concern about the co-option of AI ethics research to prevent, water-down, or negatively affect legislation and our understanding of the societal impacts of AI applications~\cite{seele2022greenwashing,young2022confronting,tafanis2022narrative}.

\noindent \textbf{This study.} The focus of this study is not to debate the benefits or harms of increased industry presence in the NLP community. 
Instead, we seek to quantify and characterize industry presence in NLP.
Using both manual annotations of the curriculum vitae (CVs) of authors who have published at a major NLP conference (namely, the 60th Annual Meeting of the Association for Computational Linguistics - ACL 2022) and automated analysis of all papers within the ACL anthology to date, we explore the industry presence in NLP using five research questions that serve as a guide for our exploration:\\
\begin{enumerate}
    \item How large is the industry presence? Who is the industry? (\S \ref{sec:industry_who})
    \vspace{-2mm}
    \item Where is the industry presence? (\S \ref{sec:industry_where})
    \vspace{-2mm}
    \item  What is industry research focused on? (\S \ref{sec:industry_what})
    \vspace{-2mm}
    \item Who does the industry work with? (\S \ref{sec:collaborations})
    \vspace{-2mm}
    \item  How well cited are industry papers? (\S \ref{sec:well_cited})
\end{enumerate}
\vspace{-2mm}
The questions for each section are expanded to provide a more fine-grained understanding of industry presence in NLP over time.
Quantifying industry presence is a vital first step to raising awareness and enabling the research community to discuss if any actions should be taken. %
The data and code for our automatic analysis are publicly available (for research purposes only) and can be found at \url{https://github.com/jpwahle/acl23-big-tech-nlp}.

\section{Related Work}
Scientometrics is a field of study which explores scientific research from a quantitative perspective~\cite{mingers2015review,leydesdorff2015scientometrics}. 
Tracing back to the mid-20th century~\cite{de1961science}, efforts in this field have focused on measuring the impact of research~\cite{garfield1979citation}, mapping science to understand relationships between fields~\cite{callon1986mapping,RuasP14}, the volume of work being published~\cite{mohammad2020nlp,lo2020s2orc}, how scientific concepts have changed over time~\cite{sharma2021drift,abdalla2022insights}, and how scientists have changed over time~\cite{mohammad2020gender,abdalla2022under}.

Focused on NLP, there has been a healthy amount of research conducted on studying the field; many researchers have shared open-sourced datasets that can be used to study the growth and change in NLP~\cite{mariani2019nlp4nlp,mohammad2020nlp,mohammad2020examining,wahle-etal-2022-d3}. 
\citet{mariani2019nlp4nlp} developed the NLP4NLP corpus, a collection of NLP articles published from 1965 to 2015 in 34 NLP conferences and journals. 
With this dataset, they provided an extensive analysis of references, citations, and authorship.

Similarly,~\citet{mohammad2020nlp} created NLP Scholar, a dataset that combined information from the ACL Anthology %
with Google Scholar, which was used to study the volume of research, the citation practices patterns in the field~\cite{mohammad2020examining}, and demographic changes in the field~\cite{mohammad2020gender}. 
\citet{wahle-etal-2022-d3} and \citet{ruas2022cs} extended NLP Scholar to include venues outside the ACL anthology with DBLP, added informative features derived from the full texts, and analyzed changes in research trends in NLP research.

Recently, a few papers have attempted to quantify the presence of industry within computer science. 
\citet{AbdallaAIES21} quantified both the number of professors at four schools who had past financial relationships with industry and the funding patterns of the biggest machine learning conferences over the past five years. 
\citet{ahmed2020democratization} created a dataset of 171,000 research papers from many computer science conferences to study industry participation rates over time. 
They found that while industry participation has been increasing over time, these collaborations have primarily remained between large companies and elite research institutions (i.e., those ranked 1-50 on QS World University Rankings\footnote{\href{https://www.topuniversities.com/qs-world-university-rankings}{www.topuniversities.com/qs-world-university-rankings}}). 

Our work stands out from previous work in multiple ways. 
Unlike most scientometrics works in the field of NLP, we are the first to focus on industry presence. 
We narrow the focus of~\citet{ahmed2020democratization} to solely NLP, but extend the scope to all conferences within the ACL anthology. 
Furthermore, our automatic analysis asks questions not explored by previous work (e.g., exploring industry-academia relationships and impact as measured by citations).

\section{Methodology}

To accurately measure industry presence in NLP research, the quality and amount of information about affiliations, funding, and employment are limited. %
In particular, few authors report %
funding or employment in a paper's acknowledgment section.

We retrieve high-quality metadata about authors 
by examining the CVs and webpages of 681 authors who published in the ACL 2022 conference.
This part of the analysis will be called \textbf{manual analysis}, %
and %
complements the automatic analysis. %

To extend our study to more venues and a larger time frame, we automatically extracted information from papers for over 34 venues in the ACL Anthology from 1965 to 2022.
This part will be referred to as the \textbf{automatic analysis}.
Although the automatic analysis cannot gather the same features as the manual analysis, it provides a historical perspective on the industry presence that can be back-tested using the manual part.
Below, we detail how we collected, processed, and annotated the data for 
automatic and manual analyses.

\subsection{Data collection}

\textbf{Automatic analysis.} 
To define the search scope of companies and universities, we use the New York Stock Exchange (NYSE) list of technology companies and an open repository of known worldwide universities.
We extract author affiliations and acknowledgment sections from the Semantic Scholar Open Research Corpus (S2ORC), which contains annotated full texts of more than 110m research papers, including the entire ACL Anthology.
To obtain metadata about the type of venue (e.g., workshop, the main event, tutorial), we use the official ACL Anthology BibTeX export.
For the topical information, we add the D3 dataset \cite{wahle-etal-2022-d3} with its Computer Science Ontology (CSO) annotations~\cite{salatino-2020-cso}.

\noindent \textbf{Manual annotation.} The manual analysis is designed to complement the automatic analysis by granting us a deeper understanding of how industry presence is manifested. 
As manual annotations are costly, we focused on a single year of a single conference, ACL 2022 (the latest flagship NLP conference). 
For each long and short paper published at ACL 2022, we randomly selected a single author from the authors list, excluding authors from being selected twice. 
For the total 701 unique authors, five of us manually searched their CVs online (the annotation instructions can be found in \Cref{appendix:Annotation-Instructions}).
To ensure we selected the right author webpage and to disambiguate between authors who shared names, we used their affiliations on the paper or Google Scholar accounts. 
For each author, we collected multiple fields, presented in \Cref{tab:manual-methodo}. 
A value of ``Unknown'' represents information we could not find. When an author had multiple entries for a single field (e.g., several internships in the same company), we listed the company name as often as it appeared. 

\begin{table}[t]
    \centering
     \resizebox{\columnwidth}{!}{
        \begin{tabular}{lrr}
        \toprule
        \textbf{Attribute} &  \textbf{Values} \\
        \midrule
        PhD graduation year                     &  Year, No, Unk. \\
        Country they currently work in           &  Country, Unk. \\
        Current company affiliations             &  Company name(s), No \\
        Title                                    &  Academic position,\\
                                                & Title in company, Unk. \\
        Past company employment                  &  Company name(s), No, Unk. \\
        Company internships                      &  Company name(s), No, Unk. \\
        Grants or awards from companies &  Company name(s), No, Unk. \\
        \bottomrule
        \end{tabular}
     }
\caption{Attributes captured using manual annotation.}
\vspace*{-2mm}
\label{tab:manual-methodo}
\end{table}

Authors were contacted by email and provided information on the study and an opportunity to withdraw from the study. 
They could also supply an updated CV (for more information on the process, see \Cref{sec:ethicalconsiderations}). 
Six authors provided their CVs, nine authors requested to be withdrawn from the study, and 11 authors could not be reached by email and were withdrawn from the study. 
In the end, information about 681 authors was used. 

\subsection{Data processing}

\noindent \textbf{Automatic analysis.} The process to reproduce our dataset can be described in five steps:
\vspace*{-2mm}
\begin{enumerate}
    \item We extracted author affiliations and acknowledgment sections from S2ORC.
    \vspace*{-2mm}
    \item We aligned the exported BibTeX from ACL Anthology with the affiliations and acknowledgments to retrieve venue information
    (e.g., Proceedings of the 16th SemEval). %
\vspace*{-2mm}
    \item We searched for technology companies and universities in the affiliations and acknowledgment entries using fuzzy matching (\Cref{sec:extraction_details}). As a proxy for Big Tech companies, we used the 100 largest technology companies by market cap according to the New York Stock Exchange (NYSE).\footnote{While NYSE is the %
    largest stock exchange in the world, 
    and includes most major technology companies (including Chinese ones), it
    may not include private companies such as OpenAI and some large non-American companies.}
\vspace*{-2mm}
    \item We aligned research topics to papers using D3 and its Computer Science Ontology (CSO) annotations using the paper's ACL anthology identifier~\cite{salatino-2020-cso}.
The CSO automatically categorizes research papers to emerging topics using syntactic and semantic analysis, based on the title and abstract with an average precision of 86.9\% (compared to 66.9\% using LDA~\cite{salatino-2020-cso}).
\vspace*{-2mm}
    \item We retrieved geolocations of affiliations, when present, using the geography API\footnote{\href{https://pypi.org/project/geograpy3/}{https://pypi.org/project/geograpy3/}}.
\end{enumerate}
\vspace*{-2mm}
\noindent In total, we processed 78,187 papers and extracted 23,606 author affiliations. Details on the standardization of affiliations %
is in \Cref{ap:faqdev}.

\noindent \textbf{Manual analysis.}
After standardization, as a sanity check, we got an additional author (not involved with the initial manual annotation) to label 20 examples (3\% of the number of authors) and measured annotator agreement.
This annotator was provided with the same set of instructions (\Cref{appendix:Annotation-Instructions}). 
The ``Exact'' column in \Cref{tab:manual-agreement} calculates what percentage of faculty annotations had the exact same list (i.e., both annotators listed the exact same set of granting companies). The ``Collapsed'' column
collapsed the list of granting companies to a simple binary (grants from industry vs. no grants from industry). Each value presents the percent agreement between annotators.
The observed agreement varied from low 0.70 to 0.90 depending on the annotated feature. 
These values are well above random agreement (0.5 for collapsed and 0.1 for exact -- a weak baseline assuming only 1 of 10 largest companies).
The majority of observed disagreements were the result of ``Unknown'' vs. ``Value'' judgment. 
That is, most errors are when one annotator finds the values and another annotator fails to do so (likely because the information is not clearly presented, different websites have different information, or a CV is not present and further investigation is needed). 
We largely used binary findings (i.e., collapsing the labels to industry vs. no industry) in our analysis.

\begin{table}[t]
    \centering
    \resizebox{0.86\columnwidth}{!}{
        \begin{tabular}{lcc}
        \toprule
        \textbf{Attribute} &  \textbf{Collapsed} & \textbf{Exact} \\
        \midrule
        URL & -- & 0.80 \\
        PhD graduation year                     &  -- & 0.75 \\
        Country they currently work in           &  -- & 0.90 \\
        Current company affiliations &  -- & 0.85 \\
        Title  &  0.90 & 0.65 \\
        Past company employment &  0.75 & 0.65 \\
        Company internships &  0.71 & 0.43 \\
        Grants or awards from companies &  0.85 & 0.54 \\
        \bottomrule
        \end{tabular}
    }
\caption{Annotator observed agreement. As multiple features could have multiple attributes (e.g., grants from multiple company names), exact measures perfect agreement, whereas collapsed indicates categorical agreement (e.g., yes past company employment or not).  }
\vspace*{-3mm}
\label{tab:manual-agreement}
\end{table}

\section{Industry Presence in NLP Research} 

\label{sec:analysis}

The manually curated data and the automatically extracted metadata allow for exploring a rich set of questions regarding the presence of Big Tech in NLP research. 
The following subsections explore a different aspect of the industry presence in NLP and rely on both manual and automated methods.

\begin{figure*}[t]
    \centering
    \includegraphics[width=0.8\textwidth]{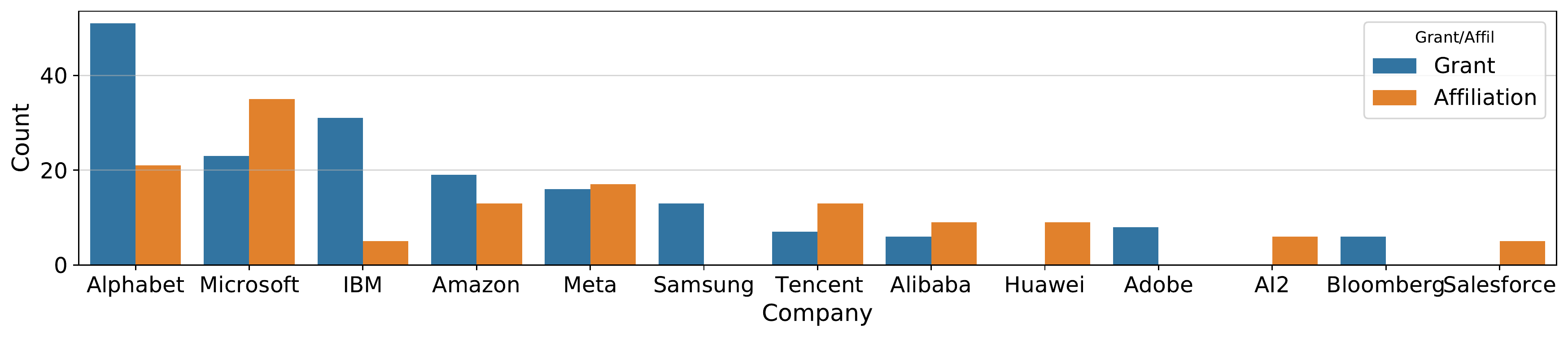}
    \vspace*{-2mm}
    \caption{The number of authors with industry affiliation (orange) and the number of grants awarded from each company (blue) evaluated in the manual analysis. Affiliations with 
    less than 5
    authors are withheld for privacy.}
    \label{fig:manual_affiliations_and_grants}
\end{figure*}

\begin{figure}[t]
    \centering
    \includegraphics[width=\columnwidth]{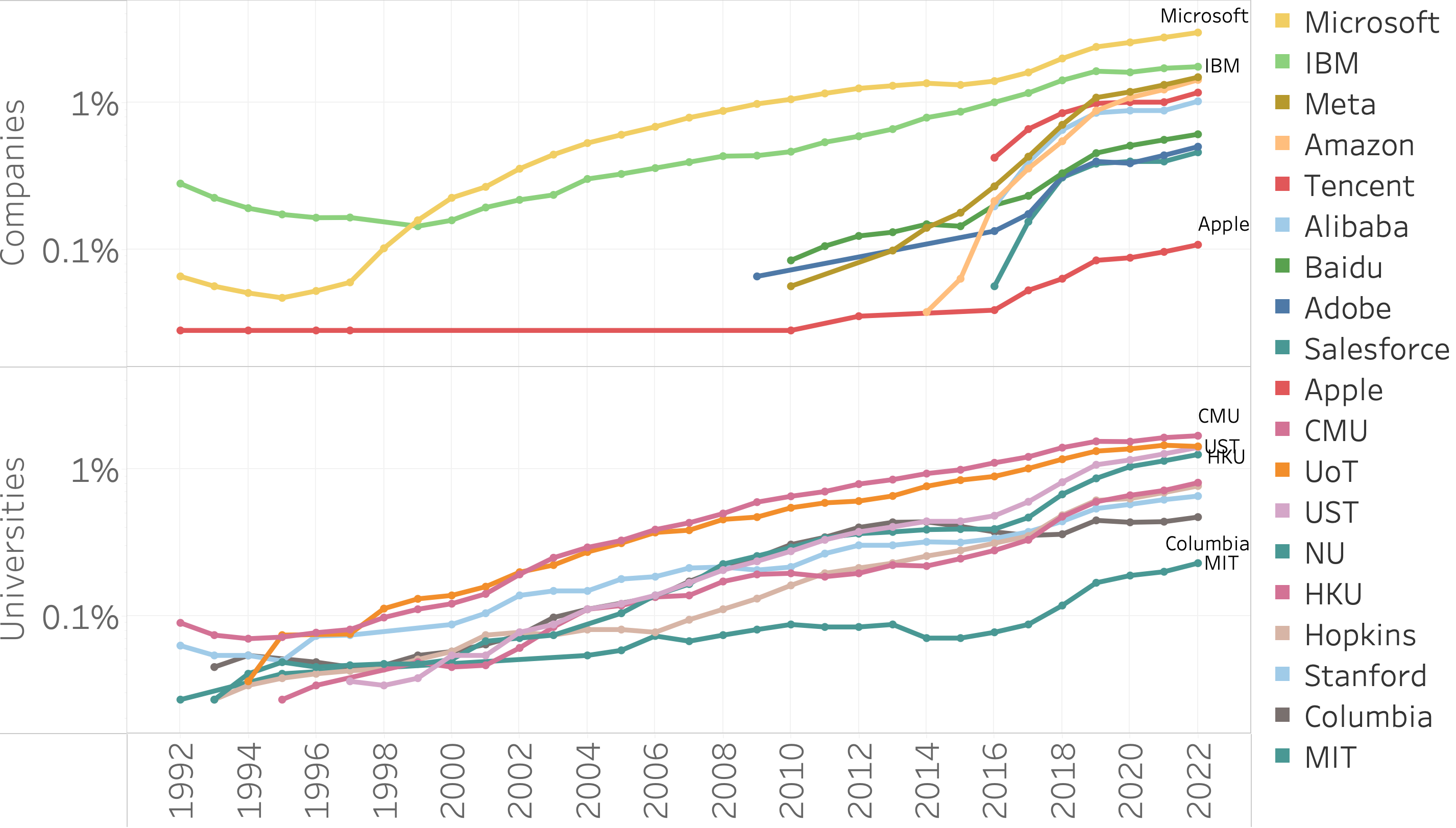}
    \caption{The relative number of papers with industry author affiliation (y-axis and in log-scale) for the top 10 companies (top) and universities (bottom) with the most papers by year. The line shows a moving average ($k=5$) of the number of papers with at least one author affiliation to that company/university divided by the total amount of papers in that year.}
    \label{fig:automatic-affiliations-over-time}
    \vspace*{-3mm}
\end{figure}

\subsection{How large is the industry presence? Who is the industry?}
\label{sec:industry_who}
\label{sec:industry_how}

Across all the papers in our corpus (1965--2022), we retrieved authors from 45 Big Tech companies (45\% of investigated companies) and 1,040 universities (37\% of investigated universities).
Microsoft had the most authors on papers (12\%), followed by IBM (8\%), Alphabet (7\%), Meta (5\%), Amazon (5\%), Alibaba (3\%), and Tencent (3\%)~\footnote{The values in parenthesis are the portions of papers with at least one author affiliation to that company over the number of all papers.}.

 \Cref{fig:overall-relative-presence} shows that the relative presence of industry has increased over time (from 1.5\% in 1995 to 14\% in 2022), with a sharp growth in the proportion of papers with at least one affiliated author in the last five years, from 5\% in 2017 to 14\% in 2022, a 180\% increase.
 This growth is largely driven by relatively recent companies such as Baidu, Meta, and Salesforce, which experienced exceptional growth over the past decade (\Cref{fig:automatic-affiliations-over-time}). 

These results are closely reflected in the manual analysis, which focused on the ACL conference in 2022. 
Of the authors who were affiliated with industry (30\% of analyzed authors), most are affiliated with Microsoft (17\%), followed by Alphabet (10\%), Meta (8\%), Amazon (6\%), and Tencent (6\%) (\Cref{fig:manual_affiliations_and_grants}).  
The proportion of authors affiliated with industry represents a 10\% increase from the numbers recorded in 2019\footnote{\href{http://acl2019pcblog.fileli.unipi.it/wp-content/uploads/2019/07/ReportACL2019ReviewingSurvey.pdf}{2019 ACL review survey}: Q48}.

The same companies are largely responsible for providing grants to faculty authors, though there is a change in their rank ordering. 
Of the authors with available information online, we observe that most grants awarded to the surveyed faculty were from Alphabet (18\% of all tracked grants), followed by IBM (11\%), Microsoft (8\%), and Amazon (7\%). 
\Cref{fig:manual_affiliations_and_grants} also illustrates how many grants and awards are provided by each company  (More on grants in \Cref{sec:industry_where}).

\subsection{Where is the industry presence?}
\label{sec:industry_where}
\subsubsection{Conferences}
The three venues with the most industry affiliations are EMNLP, ACL, and NAACL, except for 2016--2018, where COLING had more industry-affiliated papers (4\%) than NAACL (3\%).
While in 2013--2015, only 5\% of papers in ACL were affiliated with the industry, in 2019--2021, 20\% of papers are. 
For EMNLP, this trend is even stronger.
With 3\% of papers a decade ago with industry affiliations, now every fourth paper has at least one author affiliated with one of the 100 largest technology companies\footnote{For further details, Appendix \Cref{fig:automatic-venues-over-time} presents the number of industry-affiliated papers for three time spans (2013--2015, 2016--2018, 2019--2021) for the venues with the most industry-affiliated papers.}.

\begin{figure}[t]
    \centering
    \includegraphics[width=\columnwidth]{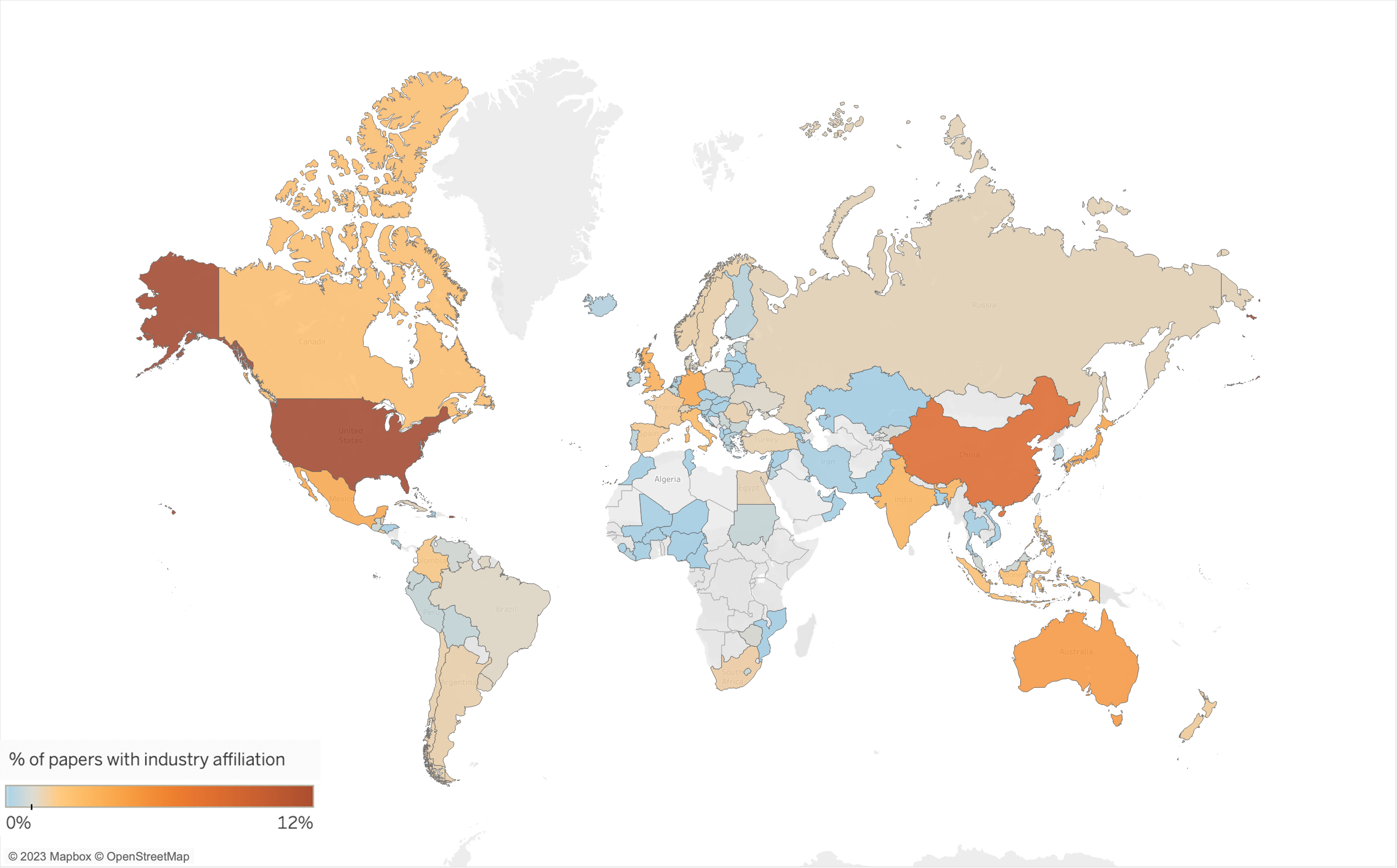}
    \caption{The percentage of papers with industry affiliation by country. Grey: no industry affiliations, light to dark blue: [0--1]\%, orange to red: [1--12]\% of papers.}
    \label{fig:automatic-world-country-map}
       \vspace*{-3mm}
\end{figure}

\subsubsection{Geographic location}
Over the entire ACL Anthology, starting from the first papers in 1965 
up to the most recent publications in 2022, %
the majority of author affiliations to companies are located in the US (29\%), followed by China (8\%), Japan (6\%), Mexico (5\%), Australia (5\%), Germany (5\%), India (4\%), Italy (4\%), and the UK (3\%). 
Further, we calculate the ratio of a paper coming from a specific country over the average of all countries to mitigate outliers. 
We provide more details on the author affiliations in \Cref{fig:automatic-world-country-map}.

The manual analysis of the ACL 2022 papers shows similar geographic trends. 
Excluding authors for whom we could not ascertain a geographic location, of all the authors with industry affiliations, 40\% come from the US, followed by China (32\%), the UK (5\%), Israel (4\%), and Canada (3\%). 

Considering the proportion of industry affiliations within geographic areas, we see that 35\% of the US authors have company affiliations. 
This is similar to the numbers observed from China (31\%) but almost double that observed in Europe (19\%).

\subsubsection{Career stages}
We explore how the relationship between researchers and industry changes as researchers progress through their careers. 
Of the authors analyzed for ACL 2022, with online profiles, a near majority of authors did not have a PhD (48\%). 
Of the 48\% of authors without a PhD, 77\% are students (which is 37\% of all authors analyzed).
The second largest groups earned their PhDs in 2018 and 2019 (each of those years represents 4\% of all authors). 
There is a steady monotonic decrease in the number of authors from then on (see \Cref{fig:manual_phd}). 
This matches what we already know about those participating in ACL Rolling Review as authors, reviewers, and editors (using seniority of reviewers as a proxy)\footnote{\href{https://www.aclweb.org/adminwiki/index.php/2022Q1_Reports:_ACL_Rolling_Review}{https://www.aclweb.org/adminwiki/index.php/2022Q1 \_Reports:\_ACL\_Rolling\_Review}, Section: ARR Today: Stats}.

\noindent \textbf{Faculty grants.} Of all analyzed faculty authors who published in ACL 2022 and had online information available, we observe that 66\% received funding (e.g., grants, research awards, etc.) from the industry. 
Stratifying this analysis by country, we observe that 72\% of US faculty who were analyzed from ACL 2022 had past funding from industry. 
This, like within country industry affiliations from the previous section,  is comparable to the Chinese faculty (69\%) but nearly double those of European faculty (38\%). 
This indicates that cultural or governance values are likely at play, impacting the likelihood for researchers to conduct research at these companies or seek industry research funding as faculty.

\noindent \textbf{Student funding and internships.} Examining opportunities open to graduate students, such as industry scholarships and internships, we observe that of the student authors with available information online, the vast majority (74\%) have either won an industry scholarship or have interned for a company.

However, when stratified by geographic origin, we see European students trending much more closely to US and Chinese students when compared to faculty or industry researchers. 
Of the US student authors, 81\% have had some financial relationship with industry. 
This is similar to Chinese student authors (75\%). 
However, we see that, unlike faculty grants, a much larger percentage of European students (65\%) have received funding from or interned for industry.

This change could indicate a shift in views towards (or a growing dependence on) industry funding in Europe amongst younger researchers. 
However, as these results are only from sampling from a single conference, more research is needed to confirm that this is part of a larger trend. 
It is also unclear whether this trend will translate into increased industry funding amongst faculty in later years.

\subsection{What is industry research focused on?}
\label{sec:industry_what}
\begin{figure}[t]
    \centering
    \includegraphics[width=0.8\columnwidth]{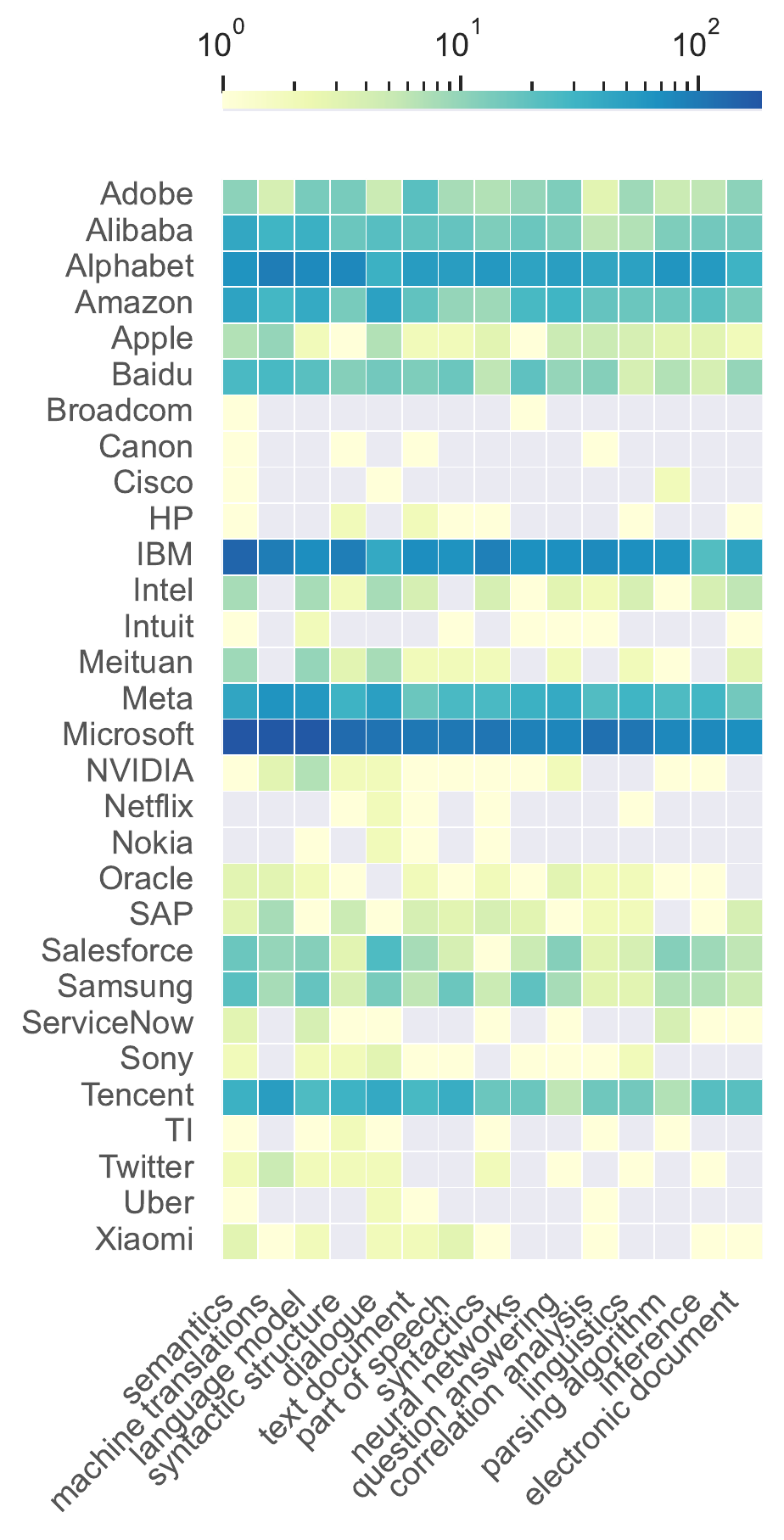}
    \caption{A heat map showing the number of papers in various areas by industry authors. The coloring from yellow to dark blue represents the number of papers (in log scale). This analysis was done on the 15 most common topics and 30 companies with the most papers. The topics are listed in descending order of their prevalence.}
    \label{fig:automatic-companies-topics-heatmap}
    \vspace*{-3mm}
\end{figure}

We analyze the 15 most %
common topics (e.g., question answering, machine translation) in papers 
by
the 30 companies with the most NLP papers.
\Cref{fig:automatic-companies-topics-heatmap} shows the results.
Some companies, such as Microsoft, publish widely across all 15 topics, whereas others, such as SAP, mainly published on word sense disambiguation, semantics, or semantic information.
Intuitively, Salesforce focuses on dialogue systems as its core business relies on customer communication.
Alphabet has a high focus on machine translation as Google Translate is one of the largest translation systems on the market.

\begin{figure}[]
    \centering
    \includegraphics[width=\columnwidth]{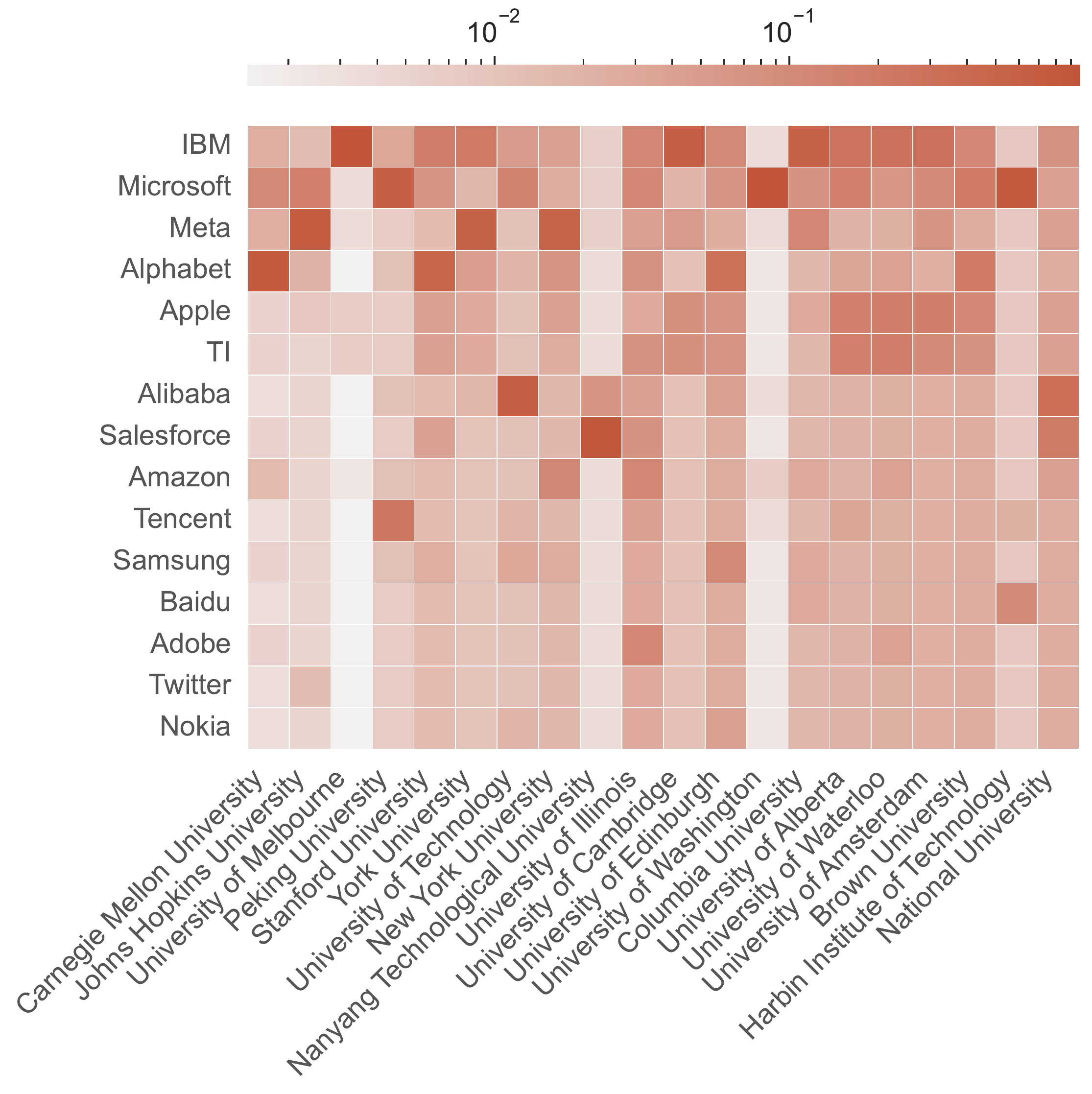}
    \caption{A heat map of the percentage of collaboration between industry and universities. The coloring (in logarithmic scale) shows the softmax normalized amount of papers (temperature $T = 2$) for which one or multiple authors had an affiliation with a company and university. Both the x-axis and y-axis are sorted for total collaborations, i.e., IBM is the company with the most papers that also have a university affiliation, and CMU 
    is the university with the most papers that also have a company affiliation.}
    \label{fig:automatic-collaboration-heatmap}
\end{figure}

\subsection{Who does the industry work with?}
\label{sec:collaborations}
We analyzed papers with multiple affiliations to 
measure the relative amount of papers in which %
affiliates from a company and university are co-authors. \Cref{fig:automatic-collaboration-heatmap} shows the heatmap.
We observed that the highest amount of joint papers were between IBM and the University of Melbourne (34\% of papers in which at least one affiliation was present). 
Some of the strongest collaborations we observe in the automatic analysis are between %
CMU and Alphabet, %
JHU and Meta, IBM and the University of Melbourne, Salesforce and 
NTU,
and the University of Washington and Microsoft. %

Many of the strong collaborations %
can be explained by geographic proximity: Microsoft is headquartered in Redmond near Seattle, the home of the University of Washington; both Nanyang Technological University and the Asia Pacific Headquarters of Salesforce are in Singapore, etc.

\subsection{How well cited are industry papers?}
\label{sec:well_cited}
\Cref{tab:citations-by-company} shows the time-weighted number of citations and h-index for the top 10 companies and universities.
Even though we rely on h-index as a proxy for impact and influence, this metric does not cover a myriad of elements in research (e.g., differentiation between fields, quality of venues, scientists' career time)~\cite{BormannD07,CostaB07}.
While Microsoft has the largest h-index (123), they have been publishing papers since 1992. 
Meta, a relatively new company, with publications starting in 2008, has already achieved an h-index of 77. 
When looking at the time-weighted number of citations, in which each paper's citations are divided by the number of years that it has been published (e.g., 100 citations for a paper published in 2012 has a time normalized value of $\frac{100}{2022 - 2012} = 10$), %
we see that Meta has a much larger median (8.00) than Microsoft (2.82). 
The top universities achieve similar time-normalized median citations as the industry. 
However, Microsoft and Alphabet have markedly higher h-indices.

A further comparison of the mean and median number of citations overall reveals the consistency of citations between papers of a single university or company (for details, see \Cref{tab:citations-by-university-full,tab:citations-by-company-full} in \Cref{sec:full_experiments_citations}). 
High medians and comparable means show that many publications have been cited similarly. High means and lower medians show that some papers have achieved a high number of citations and others did not. 
For example, Meta's mean number of citations is 113.55, while the median is 20.00, showing there have been few highly successful papers (see \Cref{tab:citations-by-company-full}). 
Although Netflix's mean number of citations is much lower (17.67), the median lies closer to it (16.00) (see \Cref{tab:citations-by-company-full}).
Another observation is that universities have a lower average h-index. 
Still, their time-weighted citation count is comparable, which can be interpreted as the industry having more one-hit successes while university research is more regular.

\begin{table}[t]
    \centering
    \resizebox{\columnwidth}{!}{
        \begin{tabular}{lrr}
        \toprule
        \textbf{Company/University} &  \textbf{T.N. Median Citns.}$^*$ &  \textbf{h-index} ($\downarrow$) \\
        \midrule
        Microsoft                        &                   2.82 &      123 \\
        Alphabet                         &                   4.50 &      104 \\
        IBM                              &                   1.60 &       94 \\
        Meta Platforms                   &                   8.00 &       77 \\
        \midrule
        Carnegie Mellon University       &                   2.14 &       95 \\
        Stanford University              &                   3.20 &       81 \\
        University of Edinburgh          &                   2.00 &       71 \\
        University of Pennsylvania       &                   1.48 &       67 \\
        University of Texas              &                   1.85 &       64 \\
        University of Cambridge          &                   2.04 &       62 \\
        \midrule
        Mean for companies and universities                   &                   1.74 &        5 \\
        \bottomrule
        \end{tabular}
    }
\caption{The median number of citations and h-index for papers in which at least one of the authors has the affiliation. We selected the top 10 affiliations by h-index. 
The rows are in decreasing order of h-index.
A more extensive list can be found in Appendix \ref{sec:full_experiments_citations}. $^*$Time normalization was performed by dividing the median by the number of years the paper was published.} 
\label{tab:citations-by-company}
\vspace*{-3mm}
\end{table}

\section{Final Considerations}

{\bf The role of industry in shaping the direction of NLP research.}
In this work, we quantified the presence of industry within the field of NLP. 
In explicitly highlighting how our field has interacted with industry in the past, we enable ourselves to make decisions for the future. 
Throughout this study, we observed that many %
researchers work in countries with vital technology industries and infrastructure, such as the US, China, Japan, and Germany. 
Additionally, these companies often collaborate with universities in these regions, providing funding and other resources for NLP research.

Advances %
made by industry %
have a significant effect on %
NLP, as well as on broader technology and society (e.g., 
widespread use of translation apps and
virtual assistants).
Overall, the presence of Big Tech companies in NLP research has contributed to the growth and advancement of the field. 
These companies provide significant funding and resources for NLP research, not least by sponsoring conferences, and their collaborations with universities and researchers have led to many important innovations and breakthroughs. 
It should be noted though, that Big Tech also benefits from their collaborations with universities and governments through tax benefits, grants, labor, recruiting, and a wide-array of open source research.

There exist fears that such tremendous infrastructure can lead to monopolies (e.g., Microsoft has an exclusive license on GPT-3, and the majority of research labs lack the infrastructure to reproduce these models.). %
Further, research that requires massive infrastructure is difficult to reproduce for most research labs.
When data, trained models, or results are not publicly shared, their transparency gets compromised.
One example is the OpenAI GPT-3 model that Microsoft exclusively licensed for \$1b in 2019, the same amount as Meta (ex. Facebook) acquired Instagram in 2012 with 80m daily active users at that time and which generates now more than \$47b in revenue per year (44\% of Facebook's revenue). %
GPT-3 was not accessible for almost a year and was then released to the general public through an API endpoint that gives access to some, but not all, functionalities of the model.
NLP and AI research %
are seen as innovation drivers for core business models --- and are valued extremely highly as such.
In contrast, the value of the research for fundamental science may be sidelined;
some symptoms of this include the widespread adoption of technologies before peer review of the underlying science --- either because the paper was simply put up on arXiv (BERT~\cite{devlin-etal-2019-bert} - for a few years) or because the technology was released only in a website (ChatGPT\footnote{\href{https://chat.openai.com/}{https://chat.openai.com/}}).
Technology companies have also had a relatively poor record of developing an inclusive work environment \cite{ScottKO17,BoagSHD22}.
Thus, there are concerns about their influence impacting the diversity of researchers in NLP. 
Several studies have shown that a diversity of participants is crucial for developing technologies that are beneficial for all, and not just those with power\footnote{\href{https://increasingdii.org/}{https://increasingdii.org/}} \cite{RaoT16,Reiche17,NielsenABE17}.  %

\noindent {\bf Recommendations.} We advocate for increased transparency in industry dealings within NLP research to help facilitate future decision-making and keep track of industry presence in the field.
The NLP community organizers should push for standardized databases of author--industry interactions, similar to the ORCID initiative~\cite{sprague2017orcid}.
We think that a centralized database of all financial relationships would go a long way to enabling the study of the effects of industry presence in our field. 
The field should also take steps to prevent the monopolization of certain kinds of research that requires huge infrastructure (e.g., the creation of publicly controlled research infrastructure possibly funded by taxes on industry). 

\section{Limitations}
\subsection{Manual Analysis}
There are a few limitations with the manual analysis. 
First, as manually annotating hundreds of CVs is a time-intensive process, our analysis only represents a single point of time for a single conference. 
While this limitation is offset by the automated analysis for certain findings (e.g., affiliation analysis), stronger conclusions regarding other findings (e.g., faculty funding through grants) cannot necessarily be drawn (though our findings are supported by previous works \cite{AbdallaAIES21}).

The findings of our manual analysis that relied solely on CV analysis may also be systematically biased by the decision of faculty and graduate students to make information about themselves publicly available. 
For many features, there were many authors with ``Unknown'', and it may be possible that this is not random but systematic (e.g., those with more industry ties choose not to publish CVs or disclose their grant history or vice-versa). 
At the same time, those who opted out of our manual annotation could have done so for reasons related to information that could be found on their CVs, and this could theoretically bias the results. 
However, we do not think that those opting out of our study had a significant impact as they represented less than 3\% of the study population.

While our annotators followed the same annotation methodology (e.g., how to find authors and annotate each author), as observed in our agreement analysis, our search engines often returned a different ordering of websites for the same query for different users. 
As annotators only had to examine the main webpage of the author, it is possible that reordering the results affected which web pages were viewed. 
However, most annotations disagreements were between a given value and ``Unknown'' rather than incorrect attribution of a feature.

\subsection{Automatic Analysis}
The extraction of texts from pdfs, as well as the extraction and matching of affiliations, typically contain more noise than human annotation, therefore, not all affiliations in papers are represented.
Particularly, older publications suffer from OCR errors and larger amounts of typos.

Furthermore, the extraction of companies in the automatic analysis is biased towards the largest 100 technology companies. 
It does not include non-publicly traded companies and non-profits such as Hugging Face or OpenAI.
In the early phases of the dataset collection, we intentionally chose a top-down approach because extracting possible company names from noisy affiliation headers requires many fuzzy text matches on named entities.

\subsection{General Limitations}
In addition to the technical limitations discussed in the previous two subsections, our research has some higher-level limitations.
Importantly, the dichotomy between university and industry is not black and white. 
We captured industry presence by looking at author affiliation and research grants. 
However, we did not look at university or department funding, which any individual researcher does not receive. 
Many universities receive (or have received) funding to sponsor their departments and, consequently, their faculty and research.
A researcher may not be directly affiliated with a company nor receive funding from any company, yet feel some pressure if their department or university is funded by industry.

This analysis is a snapshot of industry presence up until 2022. Currently, there is no automatic tool to analyze future research and industry presence or interactively set filters and generate sub-views of this study. We invite researchers to use our open-source data and code to create such an interactive system in the future.

Furthermore, we did not consider the effect of governmental or military funding on the research done at universities. 
Both government and military, like industry, have vested interests and can influence research~\cite{kistiakowsky1989military,barker2017quiet,goldfarb2008effect}. Exploration of their effect and presence over time is an area we leave to future work.

Although we quantified industry presence at multiple conferences, we did not quantify the amount of industry funding present at each conference as sponsors. 
Previous work, conference websites, and personal past experiences make us confident that most large conferences are funded, in part, by industry.

Our analysis did not stratify interaction academic-industry interactions by ethnicity, sex, or many other sensitive attributes. 
While we believe in the importance of such an analysis, the data to enable this analysis was not accessible to us: it is often not listed on websites, and the information gathered by ACL is unavailable to researchers.

Another aspect that our analysis did not touch on is that many universities are private and also require regular funding to maintain their research work.
While student registration fees cover most of the operative business, research funds typically come from state grants, federal government grants, private institutions, and industry.
We plan to trace such funding tracks of large private universities and research institutions in the future.

\section{Ethical Considerations}
\label{sec:ethicalconsiderations}

Our study does not qualify for IRB review as our study data are non-sensitive information contained in publicly available information sources. However, as part of the manual analysis, we had to create a file containing identifying information (names and website URLs to online CVs). The GDPR requires that the randomly selected participants be informed of the creation of such a file and be provided with the means to withdraw from the study and/or modify information about themselves. In compliance with the GDPR, all participants were sent an email offering them the possibility to withdraw from the study or provide an updated CV if interested. Nine participants requested the removal of their data, which we did immediately. Additionally, 11  participants could not be reached and were withdrawn from the study. Six participants sent us their updated CVs, so that we could take them into account and be more precise. 
Participants were initially reached using the emails listed in the papers, which we extracted automatically using GROBID~\cite{romary:hal-01673305}. 94 emails could not be delivered. We doubled checked the correctness of emails by looking at the first page of the published papers (and author websites if no email was present in the paper) to correct errors or find alternative emails. After correcting any mistakes, gathering alternative emails for the authors, only 11 emails returned an error. We decided not to include the participants corresponding to these emails. 
It has to be noted that no personal information regarding the individuals included was shared outside the research group, and the information was collected on a secure server and will be discarded 6 months after this research report (containing only aggregated results) has been published.

\section*{Acknowledgements}
This work was supported by the DAAD under grant No. 9187215, the Lower Saxony Ministry of Science and Culture, and the VW Foundation.

\bibliography{anthology,custom}

\appendix

\section{Appendix}
\label{sec:appendix}

\subsection{FAQ on Implementation Choices}\label{ap:faqdev}

\subsubsection*{Q1. Does the ACL Anthology provide a complete resource for NLP research?}
No. Many NLP papers are not in the ACL anthology and are published in various non-English conferences that take place locally. For example, many NLP papers are published in machine learning conferences (e.g., NeurIPS or ICML). Furthermore, many inter-disciplinary NLP works are published in journals belonging to the other discipline (e.g., clinical informatics work being published in JAMIA or JMIR).

\subsubsection*{Q2. Why do we not plot entries with less than five counts in the manual analysis?}
When presenting the results of our study, we did not want to single anyone out. Since there are only 700 papers, it may be possible to identify people who belong to groups of less than five. Five was chosen as the threshold, as it falls within commonly accepted ranges of cell size suppression used in clinical informatics research\footnote{\href{https://www.ipc.on.ca/wp-content/uploads/2017/07/ent-ices.pdf}{https://www.ipc.on.ca/wp-content/uploads/2017/07/ent-ices.pdf}}.

\subsubsection*{Q3. Why are some companies left out of the analysis?}
Including all technology companies available would result in convoluted experiments, harming our main goal. Instead of manually compiling a list of companies manually (with various selection biases), we wanted a reproducible and straightforward method to obtain the list of company names. Therefore we used the 100 largest ones by market cap according to the New York Stock Exchange (NYSE). Using the list of top 100 companies in NYSE is a reasonable choice, but we acknowledge that it too has certain selection biases (e.g., it might not include prominent companies such as HuggingFace, OpenAI, and some prominent non-US companies). We did not manually add individual company names to this list to avoid unconscious cherry picking and to allow for comparable future experiments with the same reproducible setup also for potentially other fields than technology.

\subsection{Details on Standardization}
For our analysis, we standardized the data to account for changes in company names and different ways of describing the same company (e.g., Facebook and Meta) (Table \ref{tab:standard_comps}), the various industrial titles (Table \ref{tab:standard_comppos}), and academic presence in our dataset (Table \ref{tab:standard_acadpos}).

\begin{table}[t]
   {\small
    \begin{tabular}{ll}
    \toprule
    \textbf{Names in CV} & \textbf{Std. Name} \\ 
    \midrule
    Meta, Facebook,   Meta AI                            & Meta\\                                            
    Google,   Youtube, Deepmind                          & Alphabet\\                                        
    Tencent, WeChat, Tencent (Wecht at AI)               & Tencent   \\                                     
    Microsoft, Microsft                                  & Microsoft  \\                                    
    Allen Institute for AI, AI2                          & AI2          \\                                  
    LinkedIn, Linkedin                                   & LinkedIn       \\                                       
    \bottomrule
    \end{tabular}
    }
    \caption{\label{tab:standard_comps} Company name standardization}
\end{table}

\begin{table*}[t]
\small
    \resizebox{\textwidth}{!}{
        \begin{tabular}{p{0.80\linewidth} p{0.20\linewidth}}
        \toprule
        \textbf{Position in CV} & \textbf{Standardized Position} \\
        \midrule
        Research Developer, Professional Engineer, Language Engineer, Algorithm   expert, ML and NLP engineer, Research SDE, Research and Algorithm Engineer, NLP   Engineer, Research Software Engineer, Machine Learning Engineer, Data   Scientist, Software Engineer & Junior Developer                                      \\ \\
        Senior Software Engineer, Staff Software Engineer, Senior   Engineer, Senior ML Engineer, Senior Algorithm Expert, Senior Member of   Technical Staff, Senior Staff Algorithm Engineer, Algorithm Expert, Senior   Algorithm Engineer & Senior Developer                                      \\ \\
        Senior Software Engineer (Research Scientist), Applied Scientist,   Research Consultant, Data \& Applied Scientist, Research Fellow, Applied   Research Scientist, NLP AI Researcher, Data and Applied Scientist, Machine   Learning Scientist, Applied Researcher, AI Resident, Research Council Officer,   Research Associate, Research Scientist'  & Researcher                                            \\ \\
        Senior NLP Researcher, Senior Research scientist, Senior Applied Scientist, Principal Data \& Applied Scientist, Senior Scientist, Lead Research Scientist, Senior AI Researcher, Principal Researcher, Staff Research Scientist, Lead Scientist, Chief Scientist, Senior Machine Learning Scientist, Senior Data Scientist & Senior Researcher                                     \\ \\
        CEO, Executive Officer, Chair of AI Technical Committee, Head of   Program Development, Chief Scientific Officer, Head of NLP Team, Tech Lead   Manager, Principal Applied Scientist Manager, Applied Scientist Manager, Partner   Chief Scientist, Research manager, Applied Science Manager, Technical leader,   Principal Research Scientist \& Research Manager, Partner Science Manager,   Senior Research Director, Principal Research Manager, VP of AI Research and   Applied AI, Team Leader, Research Project Manager, Senior Research Manager, Head   of Research, Senior Director, Co-founder, Director of Speech and Language Lab,   Head of lab, Director, Engineering Manager, Deputy Managing Director, Research   Director & Management\\
        \bottomrule
        \end{tabular}
    }
    \caption{\label{tab:standard_comppos} Industry position standardization}
\end{table*}

\begin{table*}[t]
\small
    \resizebox{\textwidth}{!}{
        \begin{tabular}{p{0.78\linewidth} p{0.2\linewidth}}
        \toprule
        \textbf{Position in CV} & \textbf{Standardized Position} \\
        \midrule
        Assistant Prof., Tenure-track Assistant Prof. and PhD   Advisor, Faculty, Lecturer                     & Assistant Prof.                                   \\
        Associate Prof., Senior Lecturer                                                                           & Associate Prof.                                   \\
        Prof., Full Prof., Chair Prof., Chancellor's Fellow, Reader                                       & Professor                                             \\
        Adjunct Prof.                                                                                              & Adjunct Prof.                                     \\
        Undergraduate Student                                                                                          & Undergraduate Student                                 \\
        Master Student, Masters Student, Master’s Student, Masters’   Student, MSc Student                             & Master’s Student                                      \\
        PhD Student                                                                                                                           & PhD Student                                           \\
        Postdoctoral Student, Postdoc, Postdoctoral Fellow, Post-doctoral   Research Fellow, Post-Doctoral Researcher, Postdoctoral Associate & Postdoc                        \\                        
        \bottomrule
        \end{tabular}
    }
        \caption{\label{tab:standard_acadpos} Academic position standardization}
\end{table*}

\subsection{Details on the Extraction of Affiliations}
\label{sec:extraction_details}
We searched for company names and common aliases (e.g., FAIR, Meta, Facebook) the same way as in the manual analysis using word boundary-separated regular expressions and fuzzy matching of less than two character errors.
For example, for Meta, we use the following regular expression in which each word has the ``\b'' word boundary flag.
\begin{lstlisting}
"(?=("meta|fair|facebook")){e<=2}"
\end{lstlisting}
Extracted affiliations are organized in sets per paper, i.e., when multiple authors have the same affiliation, they are counted once towards the paper being affiliated with the company.

\subsection{Dataset Versions}
We accessed data for our analysis at the following points in time.

\begin{enumerate}
    \item NYSE list of technology companies -- Date accessed/Version: 2022-10-15
    \item List of worldwide universities -- Date accessed/Version: 2022-09-04
    \item S2 Open Research Corpus -- Date accessed/Version: 2022-08-30
    \item Full ACL Anthology as BibTeX -- Date accessed/Version: 2022-10-12
    \item DBLP Discovery Dataset (D3) -- Date accessed/Version: 2022-10-01
\end{enumerate}

\subsection{Details on the Extraction of Affiliations}
\label{appendix:Annotation-Instructions}

\textbf{Each box must be “No”, “Unknown”, “Yes”, or the names of companies comma separated.}

\noindent How do we differentiate between No and Unknown? This is a bit tricky and depends on the person. For example, when looking at the website of a professor, if they don’t have a section for talks, you should put Unknown as it’s unlikely they’ve given no talks. However if they have a section for talks but none to companies it becomes a no.
On the other hand, if you have a young PhD/MSc Student who has a fully fleshed website with all their awards and such but no section for talks, you can put No talks as it’s unlikely they’ve had them and forgot to put them.
Likewise for grants. It’s unlikely that a senior professor with 20 years experience has won 0 grants. So, if they don’t list ANY grants we’d put “Unknown” instead of No. However, grad students/industry workers (if they were never faculty) likely have not won grants. Here you have lee-way which will be standardized later to putting “No” if they have a full CV available but no grants section, or “Unknown” if no full CV. Either way, for these two groups standardization will adjust them.

\subsubsection{Annotation Instructions}
Below are the annotation instructions for each characteristic of interest.\\

\noindent \underline{\textbf{URL}}

\noindent \textbf{Description:} The URL for their personal/academic website.

\noindent \textbf{Annotation Instructions:} Goal is to find their academic page. How to find it depends on the person. We start by googling “name” + “nlp”. If a likely positive result shows up, we investigate by looking at their publications on the website and comparing to paper title in the sheet.  If their list of publications is not up to date or you’re not confident, looking at their affiliation on the paper can help you disambiguate. If googling “name” + “nlp” doesn’t work, you can then try “name” + “institution”. You can also try to see if their google scholar profile links to a web-page. If you can’t find a URL put “Unknown”. \\

\noindent \underline{\textbf{PhD Year}}

\noindent \textbf{Description:} Year they graduate PhD.

\noindent \textbf{Annotation Instructions:} Usually in CVs, Bios, etc. If you can’t find it put “Unknown”. \\

\noindent \underline{\textbf{Country of work/affiliation}}

\noindent \textbf{Description:} The country of their current affiliation.

\noindent \textbf{Annotation Instructions:} E.g., A PhD student at UofT is “Canada” (even if they are recently immigrated from China, for example). Sometimes this is not clear (e.g., no webpage + unclear affiliations from). If they put a company (e.g., Microsoft), but don’t specify where (and no other information helps you see), put “Unknown” as Microsoft has offices in multiple places. Hong Kong will be standardized to China. \\

\noindent \underline{\textbf{Is affiliated to a Company?}}

\noindent \textbf{Description:} Are they presently affiliated with a company?

\noindent \textbf{Annotation Instructions:} 
\begin{itemize}
    \item If they are in Industry, then put the name of company they work for. 
    \item If they are in academia and have no dual affiliation/current job at the same time, put “No”.
    \item If they are in academia AND industry (whether an official dual affiliation or working in two places at once) note: “DUAL (Place1, Place2)” This will be dealt with during standardization.
    \item If it’s not clear where they currently are, pull it from the paper.
\end{itemize} 

\noindent \underline{\textbf{Role}}

\noindent \textbf{Description:} What is the role at their place(s) of affiliation?

\noindent \textbf{Annotation Instructions:} For each affiliation from the previous column, put their current role using comma to separate roles. Sometimes, if you only have an OpenReview profile they will not state their role. In this case put “Unknown”. We do not distinguish between Endowed Professorships (e.g., “The NameA NameB Professor of Subject” would be reduced to “Professor”). \\

\noindent \underline{\textbf{Grant(s) from company}}

\noindent \textbf{Description:} Has the individual received funding for their research? 

\noindent \textbf{Annotation Instructions:} 
\begin{itemize}
    \item For Professors: This will most generally appear in the form of grants + research awards + general awards. If they have won a grant or received some research funding from a company, put the company name separated with commas. E.g., “Google, Amazon, Microsoft, Google” would be a professor that has received 2 Google grants, 1 Amazon, 1 Microsoft.
    \item For students: If they’ve received grants (quite unlikely), follow the instructions above. Otherwise put “No” or “Unknown”, this will be standardized later.
    \item For industry: If they’ve received grants (quite unlikely), follow the instructions above. Otherwise put “No” or “Unknown”, this will be standardized later.
\end{itemize} 

\noindent \underline{\textbf{Visiting Researcher}}

\noindent \textbf{Description:} Have they had a visiting researcher role in industry?

\noindent \textbf{Annotation Instructions:} 
\begin{itemize}
    \item For Professors: This is different from past work at a company. A visiting researcher role is often a short (often 1 year) stint at a company after which they return to the same academic position. If they put their work history and you can’t see them use this term put “No”. If they don’t put their work history, only then put “Unknown”.
    \item For students/industry: Quite unlikely (as this would just be an internship). If their CV is fully there and they use the term, but you see internships you can put “No”. If no CV/no past work history, you can put “Unknown” (this will be dealt with during standardization).
\end{itemize} 

\noindent \underline{\textbf{Graduate Fundings from companies:}}

\noindent \textbf{Description:} Has the individual received graduate funding/awards during their PhD? 

\noindent \textbf{Annotation Instructions:} This only deals with things during graduate education. If you see it, put the company names separated by commas. Often, professors and those in industry do not put things from their graduate education. If you feel they are omitting information from their time in graduate school (e.g., no mention of any awards/funds/internships during grad school), put “Unknown”. If they have complete information during their graduate education but nothing from companies, put “No”. \\

\noindent \underline{\textbf{Internship in a company}}

\noindent \textbf{Description:} Has the individual interned for companies?

\noindent \textbf{Annotation Instructions:} This only deals with internships during education (otherwise it’d be employment). If you see it, put the company names separated by commas. Often, professors and those in industry do not put things from their graduate time. If you feel they are omitting information from their time in graduate school (e.g., no mention of any awards/funds/internships during grad school), put “Unknown”. If they have complete information during their graduate education but no internships, put “No”. \\

\noindent \underline{\textbf{Past industry work/research (any other types)}}

\noindent \textbf{Description:} Catch-all industry financial connection category

\noindent \textbf{Annotation Instructions:} This is the catchall category. Have they received financial compensation in some way from a company. This includes past work experience (not internships), working as an advisor/consultant, being the founder but no longer part of, etc. If there is no employment history available for you to view, put “Unknown”. Otherwise “No”, or company names comma-separated.  \\

\noindent \underline{\textbf{Comments}}

\noindent \textbf{Description:} N/A

\noindent \textbf{Annotation Instructions:} If you have any comments about the person. Either some information you had a hard time processing or something you wanted to point out. \\

\subsection{Additional Details of our Experiments}
In this section, we provide more details about our experiments in \Cref{sec:analysis}.

\subsubsection{Venues and Tracks}
\begin{figure*}[h]
    \centering
    \includegraphics[width=\textwidth]{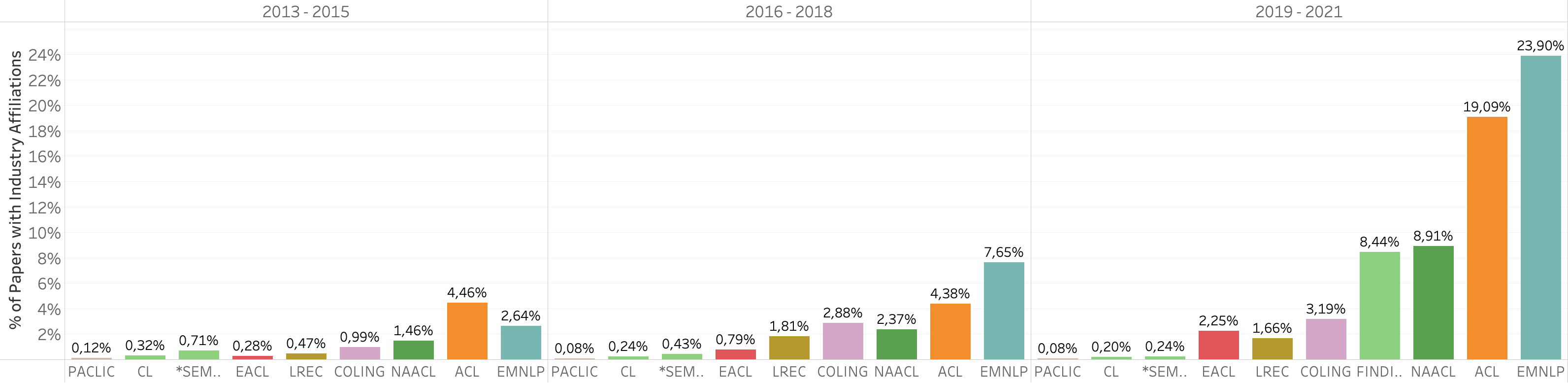}
    \caption{The relative number of papers by year with industry author affiliations by venues. We select the venues with the most industry affiliations per year. We grouped years into three consecutive bins to ensure that each event happened at least once.}
    \label{fig:automatic-venues-over-time}
\end{figure*}

\Cref{fig:automatic-venues-over-time} shows the number of papers by year with industry author affiliations by venues.

\subsubsection{Manual: Career Stage Analysis}
\begin{figure*}[h]
    \centering
    \includegraphics[width=\textwidth]{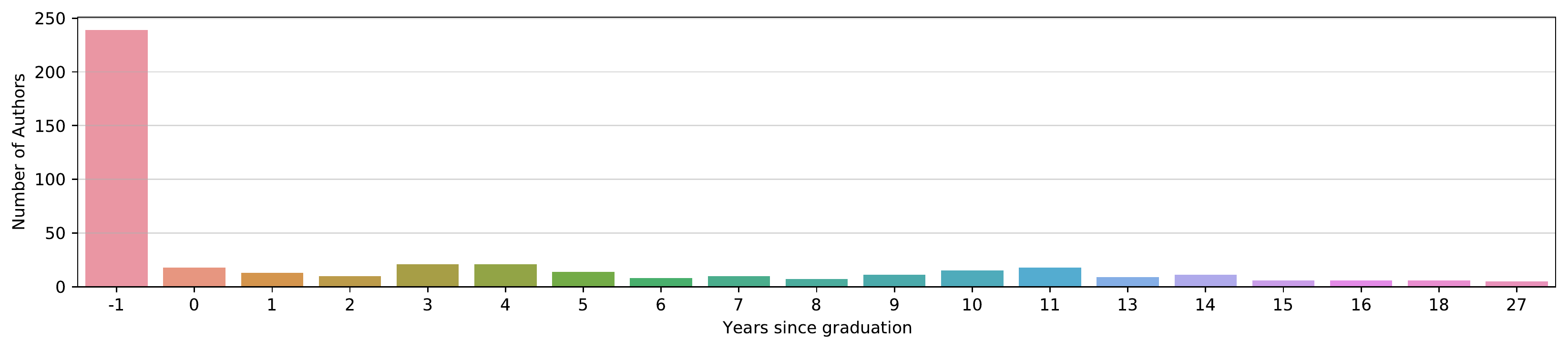}
    \caption{Author seniority measured by the number of authors (y-axis) that have a PhD for at least $n$ years (x-axis). `-1' represents an author who has not yet earned a PhD. For privacy purposes, we do not plot any PhD year with less than five authors. We do not plot authors for whom we could not confidently assign a year for their PhD (i.e., labeled `Unknown' -- 185 individuals).}
    \label{fig:manual_phd}
\end{figure*}

Figure \ref{fig:manual_phd} presents author seniority as measured by years since PhD. 

\subsubsection{Manual: Geographic Analysis}
\begin{figure*}[h]
    \centering
    \includegraphics[width=\textwidth]{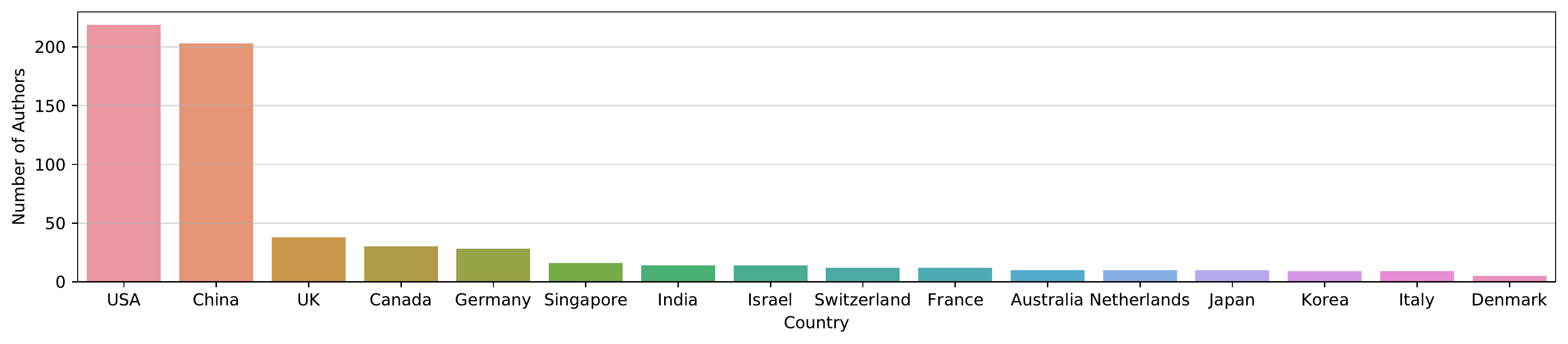}
    \caption{The number of authors per country for the 682 authors from ACL 2022 who were manually examined. An author is assigned to a country if their institution is found within the said country (e.g., Stanford is classified as USA). For privacy purposes, we do not plot any country with fewer than 5 authors. We do not plot authors for whom we could not reliably assign a country of origin (i.e., labeled `Unknown' -- 32 individuals).}
    \label{fig:manual_country}
\end{figure*}

Figure \ref{fig:manual_country} presents the country of affiliation for each author. 

\subsubsection{Citation Analysis}
\Cref{tab:citations-by-university-full,tab:citations-by-company-full} show the mean and the median number of citations as well as h-index for papers in which at least one of the authors with a university and industry affiliation respectively.
\label{sec:full_experiments_citations}
\begin{table*}[h]
    \centering
    \resizebox{\textwidth}{!}{
        \begin{tabular}{lrrrrr}
        \toprule
        University &   Mean &  Median &  Time Norm.    Mean$^*$ &  Time Norm. Median$^*$ &  h-index ($\downarrow$) \\
        \midrule
        Carnegie Mellon University                     &  65.06 &   15.00 &                    9.33 &                   2.14 &                   95.00 \\
        Stanford University                            & 129.73 &   21.00 &                   16.24 &                   3.20 &                   81.00 \\
        University of Edinburgh                        &  54.34 &   18.00 &                    6.03 &                   2.00 &                   71.00 \\
        University of Pennsylvania                     &  99.34 &   12.00 &                    6.38 &                   1.48 &                   67.00 \\
        University of Texas                            &  36.16 &   11.00 &                    3.97 &                   1.85 &                   64.00 \\
        University of Cambridge                        &  45.19 &   13.00 &                    6.12 &                   2.04 &                   62.00 \\
        University of Technology                       &  25.45 &    6.00 &                    3.23 &                   1.00 &                   60.00 \\
        Johns Hopkins University                       &  54.09 &   15.50 &                    7.74 &                   2.92 &                   59.00 \\
        University of Washington                       &  59.19 &   17.00 &                   11.35 &                   3.62 &                   58.00 \\
        Columbia University                            &  64.47 &   17.00 &                    6.48 &                   2.50 &                   58.00 \\
        University of Illinois                         &  38.89 &   14.00 &                    5.60 &                   2.18 &                   58.00 \\
        York University                                &  73.40 &   16.00 &                   10.21 &                   2.00 &                   56.00 \\
        Institute of Science and Technology            &  30.38 &    7.00 &                    2.84 &                   0.87 &                   55.00 \\
        New York University                            &  97.89 &   12.00 &                   14.21 &                   1.94 &                   50.00 \\
        Brown University                               &  89.70 &   36.00 &                    7.25 &                   2.38 &                   50.00 \\
        University of Melbourne                        &  30.58 &    9.00 &                    4.44 &                   1.50 &                   48.00 \\
        University of Sheffield                        &  33.90 &   12.00 &                    4.06 &                   1.51 &                   45.00 \\
        University of Southern California              &  76.89 &   13.00 &                    6.39 &                   2.17 &                   43.00 \\
        City University                                &  20.77 &    8.00 &                    2.43 &                   1.00 &                   43.00 \\
        National University                            &  17.15 &    4.00 &                    3.25 &                   1.00 &                   42.00 \\
        University of Trento                           &  35.51 &   11.00 &                    4.27 &                   1.79 &                   41.00 \\
        Cornell University                             &  53.74 &   18.00 &                    8.95 &                   3.00 &                   40.00 \\
        Macquarie University                           &  38.51 &   18.50 &                    3.96 &                   2.00 &                   37.00 \\
        University of Amsterdam                        &  30.87 &   10.00 &                    6.28 &                   2.00 &                   37.00 \\
        Peking University                              &  27.27 &    8.00 &                    4.61 &                   2.00 &                   36.00 \\
        Ohio State University                          &  35.52 &    8.00 &                    5.05 &                   1.00 &                   36.00 \\
        Chinese University of Hong Kong                &  26.52 &    9.50 &                    5.25 &                   2.23 &                   35.00 \\
        Hong Kong Polytechnic University               &  18.61 &    4.00 &                    2.65 &                   0.76 &                   32.00 \\
        Brandeis University                            &  73.00 &    9.00 &                    5.25 &                   1.18 &                   32.00 \\
        Massachusetts Institute of Technology          &  75.64 &   17.50 &                    9.56 &                   3.56 &                   32.00 \\
        University of Groningen                        &  22.64 &    9.00 &                    3.48 &                   1.33 &                   31.00 \\
        Harbin Institute of Technology                 &  45.94 &   10.00 &                    7.01 &                   1.65 &                   31.00 \\
        Toyota Technological Institute                 &  58.66 &   14.50 &                    9.77 &                   3.58 &                   30.00 \\
        University of Science and Technology           &  20.79 &    4.00 &                    3.27 &                   0.72 &                   30.00 \\
        University of Pittsburgh                       &  37.50 &   14.00 &                    3.60 &                   1.67 &                   30.00 \\
        University of Wolverhampton                    &  36.73 &    7.00 &                    6.70 &                   1.25 &                   29.00 \\
        Harvard University                             &  41.08 &   17.00 &                    4.79 &                   1.73 &                   29.00 \\
        Georgia Institute of Technology                &  46.66 &   10.00 &                    9.25 &                   3.00 &                   28.00 \\
        Hong Kong University of Science and Technology &  24.63 &   12.00 &                    6.65 &                   4.12 &                   28.00 \\
        Dublin City University                         &  21.43 &   10.00 &                    2.56 &                   1.26 &                   27.00 \\
        \midrule
        Mean$^\dagger$                                 &  19.21 &   10.61 &                    2.85 &                   1.74 &                    5.48 \\
        \bottomrule
        \end{tabular}
    }
\caption{The mean and the median number of citations as well as h-index for papers in which at least one of the authors has a university affiliation. We selected the top 40 affiliations by h-index. $^*$Time normalization was performed by dividing the mean/median by the number of years the paper was published. $^\dagger$The mean over all companies and universities (also ones not listed here).}
\label{tab:citations-by-university-full}
\end{table*}

\begin{table*}[p]
    \centering
    \small
        \begin{tabular}{lrrrrr}
        \toprule
        Company &   Mean &  Median &  Time Norm.    Mean$^*$ &  Time Norm. Median$^*$ &  h-index ($\downarrow$) \\
        \midrule
        Microsoft                        &  61.32 &   16.00 &                    9.05 &                   2.82 &                  123.00 \\
        Alphabet                         &  81.62 &   21.00 &                   15.48 &                   4.50 &                  104.00 \\
        IBM                              &  74.22 &   11.00 &                    6.53 &                   1.60 &                   94.00 \\
        Meta Platforms                   & 113.55 &   20.00 &                   25.47 &                   8.00 &                   77.00 \\
        Tencent                          &  25.91 &    9.00 &                    6.52 &                   3.50 &                   42.00 \\
        Baidu                            &  32.79 &    8.00 &                    7.01 &                   2.50 &                   36.00 \\
        Amazon                           &  21.19 &    4.00 &                    5.64 &                   1.50 &                   35.00 \\
        Alibaba                          &  16.16 &    6.00 &                    4.92 &                   2.50 &                   30.00 \\
        Salesforce                       &  36.67 &   12.00 &                   11.14 &                   5.00 &                   28.00 \\
        Adobe                            &  13.81 &    6.00 &                    4.13 &                   2.00 &                   22.00 \\
        Samsung                          &  13.94 &    3.00 &                    3.27 &                   1.00 &                   18.00 \\
        Apple                            &  46.92 &    9.00 &                    7.09 &                   2.00 &                   15.00 \\
        SAP                              &  39.42 &    9.50 &                    6.99 &                   1.52 &                   15.00 \\
        Intel                            &  17.05 &    8.00 &                    6.78 &                   2.50 &                   13.00 \\
        Sony                             &  73.39 &   12.00 &                    8.80 &                   0.64 &                   12.00 \\
        Meituan                          &   8.44 &    2.00 &                    4.74 &                   1.17 &                    8.00 \\
        Nokia                            &  15.11 &   13.00 &                    1.32 &                   1.43 &                    7.00 \\
        NVIDIA                           &  10.32 &    3.00 &                    4.14 &                   2.00 &                    7.00 \\
        Oracle                           &   5.54 &    1.00 &                    1.85 &                   0.50 &                    6.00 \\
        Xiaomi                           &   6.50 &    4.00 &                    2.62 &                   2.00 &                    6.00 \\
        Intuit                           &  45.14 &   34.00 &                   10.84 &                   6.80 &                    5.00 \\
        Twitter                          &   5.08 &    3.50 &                    1.75 &                   1.25 &                    5.00 \\
        HP                               &  11.67 &    5.50 &                    0.95 &                   0.48 &                    4.00 \\
        Block                            &  20.57 &    5.00 &                    1.25 &                   0.31 &                    4.00 \\
        ServiceNow                       &   7.75 &    5.00 &                    4.62 &                   4.50 &                    4.00 \\
        Canon                            &  12.00 &   15.50 &                    1.33 &                   0.65 &                    3.00 \\
        Texas Instruments                &  51.29 &    0.00 &                    1.58 &                   0.00 &                    3.00 \\
        Uber                             &  10.00 &    9.00 &                    2.56 &                   2.25 &                    3.00 \\
        Cisco                            &   5.00 &    6.00 &                    1.30 &                   1.40 &                    2.00 \\
        Airbnb                           &  15.50 &   15.50 &                    2.72 &                   2.72 &                    2.00 \\
        NetEase                          &   3.50 &    1.50 &                    1.02 &                   0.67 &                    2.00 \\
        Netflix                          &  17.67 &   16.00 &                    3.53 &                   3.20 &                    2.00 \\
        Broadcom                         &   3.00 &    4.00 &                    1.28 &                   1.33 &                    2.00 \\
        Autodesk                         &  10.00 &    2.00 &                    0.86 &                   0.50 &                    2.00 \\
        PayPal                           &   5.00 &    5.00 &                    1.67 &                   1.67 &                    1.00 \\
        Tesla                            &   8.00 &    8.00 &                    0.89 &                   0.89 &                    1.00 \\
        \midrule
        Mean$^\dagger$                   &  21.15 &    7.10 &                    4.09 &                   1.77 &                   16.62 \\
        \bottomrule
        \end{tabular}
\caption{The mean and the median number of citations as well as h-index for papers in which at least one of the authors has an industry affiliation. We selected the top 10 affiliations by h-index. $^*$Time normalization was performed by dividing the mean/median by the number of years the paper was published. $^\dagger$The mean over all companies and universities (also ones not listed here).}
\label{tab:citations-by-company-full}
\end{table*}

\cleardoublepage
\onecolumn
\hypertarget{annotation}{}
\citationtitle

\onlineversion{https://aclanthology.org/2023.acl-long.734/}
\begin{bibtexannotation}
@inproceedings{abdalla-etal-2023-elephant,
    title = "The Elephant in the Room: Analyzing the Presence of Big Tech in Natural Language Processing Research",
    author = "Abdalla, Mohamed  and
      Wahle, Jan Philip  and
      Ruas, Terry  and
      N{\'e}v{\'e}ol, Aur{\'e}lie  and
      Ducel, Fanny  and
      Mohammad, Saif  and
      Fort, Karen",
    editor = "Rogers, Anna  and
      Boyd-Graber, Jordan  and
      Okazaki, Naoaki",
    booktitle = "Proceedings of the 61st Annual Meeting of the Association for Computational Linguistics (Volume 1: Long Papers)",
    month = jul,
    year = "2023",
    address = "Toronto, Canada",
    publisher = "Association for Computational Linguistics",
    url = "https://aclanthology.org/2023.acl-long.734",
    doi = "10.18653/v1/2023.acl-long.734",
    pages = "13141--13160"
}\end{bibtexannotation}

\end{document}